\documentclass[runningheads]{llncs}
\usepackage[T1]{fontenc}
\usepackage{booktabs}
\usepackage[misc]{ifsym}
\usepackage{graphicx,verbatim}
\usepackage{amsmath}
\usepackage{amssymb}
\usepackage{algorithm}
\usepackage{algorithmic}
\usepackage{graphicx}
\usepackage{subcaption}
\usepackage{url}
\usepackage{pifont}
\usepackage{xcolor}
\usepackage{threeparttable}
\usepackage{multirow}
\usepackage{float}

\usepackage{mwe}

\begin{document}

\title{Shattering the Shortcut: A Topology-Regularized Benchmark for Multi-hop Medical Reasoning in LLMs}


\author{Xing Zi\inst{1} \and
Xinying Zhou\inst{1} \and
Jinghao Xiao\inst{1} \and
Catarina Moreira\inst{1} \and
Mukesh Prasad\inst{1}}

\authorrunning{X. Zi et al.}

\institute{University of Technology Sydney, Australia\\
\email{\{xing.zi-1, jinghao.xiao, catarina.pintomoreira, mukesh.prasad\}@uts.edu.au},\\
\email{roxyzhou210@gmail.com}}



\maketitle              

\begin{abstract}
While Large Language Models (LLMs) achieve expert-level performance on standard medical benchmarks through single-hop factual recall, they severely struggle with the complex, multi-hop diagnostic reasoning required in real-world clinical settings. A primary obstacle is "shortcut learning", where models exploit highly connected, generic hub nodes (e.g., "inflammation") in knowledge graphs to bypass authentic micro-pathological cascades. To address this, we introduce ShatterMed-QA, a bilingual benchmark of 10,558 multi-hop clinical questions designed to rigorously evaluate deep diagnostic reasoning. Our framework constructs a topology-regularized medical Knowledge Graph using a novel $k$-Shattering algorithm, which physically prunes generic hubs to explicitly sever logical shortcuts. We synthesize the evaluation vignettes by applying implicit bridge entity masking and topology-driven hard negative sampling, forcing models to navigate biologically plausible distractors without relying on superficial elimination. Comprehensive evaluations of 21 LLMs expose a systemic vulnerability to shortcut learning: frontier models fall into topology-driven distractor traps at a staggering rate (e.g., reaching a 53\% Hard Negative Error rate, far exceeding the 33\% random baseline). Crucially, restoring the masked evidence via Retrieval-Augmented Generation (RAG) triggers substantial recoveries (up to nearly 70\% Reasoning Recovery Rate), validating ShatterMed-QA's structural fidelity and proving that current models suffer from severe topological knowledge gaps rather than pure reasoning engine failures. Explore the dataset, interactive examples, and full leaderboards at our project website: \url{https://shattermed-qa-web.vercel.app/}

\end{abstract}

\section{Introduction}
The application of Large Language Models (LLMs) in the biomedical domain has significantly advanced clinical informatics~\cite{thirunavukarasu2023large}. Early evaluations primarily tested these models on their ability to memorize and retrieve static knowledge, using foundational medical question-answering (QA) benchmarks like MedQA~\cite{jin2021disease}, PubMedQA~\cite{jin2019pubmedqa}, and MultiMedQA~\cite{singhal2025toward}. While state-of-the-art models perform well on these standardized exams, the core requirements for medical AI are shifting~\cite{singhal2025toward}. As these systems transition from theoretical concepts to real-world diagnostic tools, they must go beyond single-hop factual recall~\cite{kim2025biohopr}~\cite{zuo2025medxpertqa}. Real-world clinical diagnosis requires rigorous, multi-hop reasoning that connects patient presentations to underlying pathophysiological mechanisms, laboratory results, and treatment plans~\cite{kim2025biohopr}~\cite{wu2025medreason}~\cite{zuo2025medxpertqa}.

Despite their success on straightforward tasks, current advanced LLMs struggle with deep, multi-layered clinical reasoning~\cite{zuo2025medxpertqa}. A major obstacle is \textit{shortcut learning}, where models rely on superficial correlations rather than true causal mechanisms. In medical knowledge graphs (KGs), this often means models exploit highly connected "hub nodes" (such as generic terms like "blood" or "inflammation") to guess answers, bypassing the actual micro-pathological pathways~\cite{obraczka2022fast,jiang2023path}. Furthermore, existing datasets mostly test explicit factual retrieval where intermediate steps are clearly stated. Conversely, authentic clinical environments necessitate \textit{implicit reasoning}, defined as the capability to deduce unstated transitional steps, or bridge entities, that link a symptom directly to a disease. When forced to perform this implicit reasoning, even the most advanced models show a significant drop in accuracy~\cite{kim2025biohopr}. Finally, current automated methods for generating complex QA datasets often rely on unconstrained black-box models, which can introduce hallucinations and lack traceability~\cite{liu2025med}. Without a clear way to track how the data was generated, it is difficult to guarantee clinical safety or verify the exact source evidence of a model's reasoning path~\cite{wu2025medreason}.

To overcome these critical limitations, we propose an end-to-end framework for medical Knowledge Graph (KG) construction and multi-hop benchmark synthesis. We introduce \textbf{ShatterMed-QA}, a rigorous dataset designed to evaluate deep diagnostic reasoning rather than superficial recall. Our methodology tackles shortcut learning at its root by employing a novel topology-regularization technique ($k$-Shattering) that physically prunes generic hub nodes, forcing models to navigate true micro-pathological cascades. Furthermore, our generation pipeline enforces strict traceability by anchoring every synthesized question to explicit medical source evidence. By integrating implicit bridge entity masking and topology-driven hard negative sampling, we challenge models with highly deceptive, biologically plausible distractors.

The primary contributions of this work are summarized as follows:
\begin{itemize}
\item \textbf{An End-to-End Data Synthesis Framework:} We propose a novel, automated pipeline for generating complex reasoning tasks. By integrating topology-regularized KG construction ($k$-Shattering) with constrained QA synthesis (implicit masking and hard negative sampling), this framework systematically eradicates shortcut learning and generative hallucinations.
\item \textbf{The ShatterMed-QA Benchmark:} Utilizing our framework, we introduce a bilingual (English and Chinese) dataset of 10,558 multi-hop clinical QA pairs. This includes a rigorously physician-vetted \textit{Golden Subset} of 264 highly complex diagnostic vignettes, establishing a pristine evaluation ground for frontier LLMs.
\item \textbf{Comprehensive Evaluation \& Clinical Insights:} We extensively benchmark 21 state-of-the-art LLMs, exposing critical multi-hop reasoning deficits, particularly within current domain-specific models. Crucially, we empirically demonstrate that Retrieval-Augmented Generation (RAG) resolves these failures, validating our benchmark's structural fidelity and offering new diagnostic insights into model capabilities.
\end{itemize}

\section{Related Work}

\subsection{Foundational Medical QA Benchmarks}
Medical large language model (LLM) evaluation has historically relied on datasets from professional licensing exams and biomedical literature~\cite{thirunavukarasu2023large}. MedQA~\cite{jin2021disease} established a baseline using multiple-choice questions from US (USMLE), Mainland China, and Taiwan exams. Similarly, PubMedQA~\cite{jin2019pubmedqa} shifted focus to scientific reasoning, tasking models with answering questions using PubMed abstracts. To standardize evaluation, MultiMedQA aggregated existing datasets alongside HealthSearchQA, and creating a comprehensive framework for models like Med-PaLM~\cite{singhal2025toward}. While providing robust proxies for clinical memorization, these foundational benchmarks rely heavily on static textbook knowledge and often fail to assess deep, multi-hop pathological reasoning.

\subsection{Cross-Lingual and Global Health Generalization}
As LLMs achieved expert performance on Western exams, evaluating their generalization across diverse cultural and linguistic contexts became urgent. AfriMed-QA~\cite{olatunji2024afrimed} addresses Global South underrepresentation with a pan-African benchmark sourced from 16 countries, testing epidemiological reasoning and adherence to local guidelines. Concurrently, MMedBench~\cite{qiu2024towards} introduced a multilingual QA dataset across six languages. It uniquely requires models to generate high-quality rationales alongside answers to ensure transparent explainability.

\subsection{Knowledge Graph Integration and Synthetic Benchmark Construction}
Recent dataset construction has shifted towards synthetic evolution and neuro-symbolic engineering to challenge frontier LLMs. The UltraMedical pipeline uses the InstructEvol methodology to artificially inflate semantic complexity, producing advanced synthetic datasets like MedQA-Evol~\cite{zhang2024ultramedical}. To explicitly enforce logical rigidity and evaluate multi-hop inference, researchers increasingly utilize structured Knowledge Graphs (KGs)~\cite{kim2025biohopr,wu2025medreason,liu2025med}. For instance, BioHopR extracts 1-hop and 2-hop reasoning tasks directly from the PrimeKG ontology to test the resolution of implicit intermediate entities~\cite{kim2025biohopr}. MedReason~\cite{wu2025medreason} maps unstructured clinical QA pairs to a KG to elicit verifiable ''thinking paths'', ensuring rationales are medically factual rather than parametric guesswork. Furthermore, frameworks like Med-CRAFT use deterministic KG traversal to ensure strict logical provenance and eliminate generative hallucinations during dataset creation~\cite{liu2025med}. Building upon this trajectory, our topology-regularized framework ($k$-shattering) actively mitigates shortcut learning and spurious correlations, establishing a new paradigm for multi-hop clinical evaluation.
\begin{figure*}[t]
    \centering
    \includegraphics[width=1.0\textwidth]{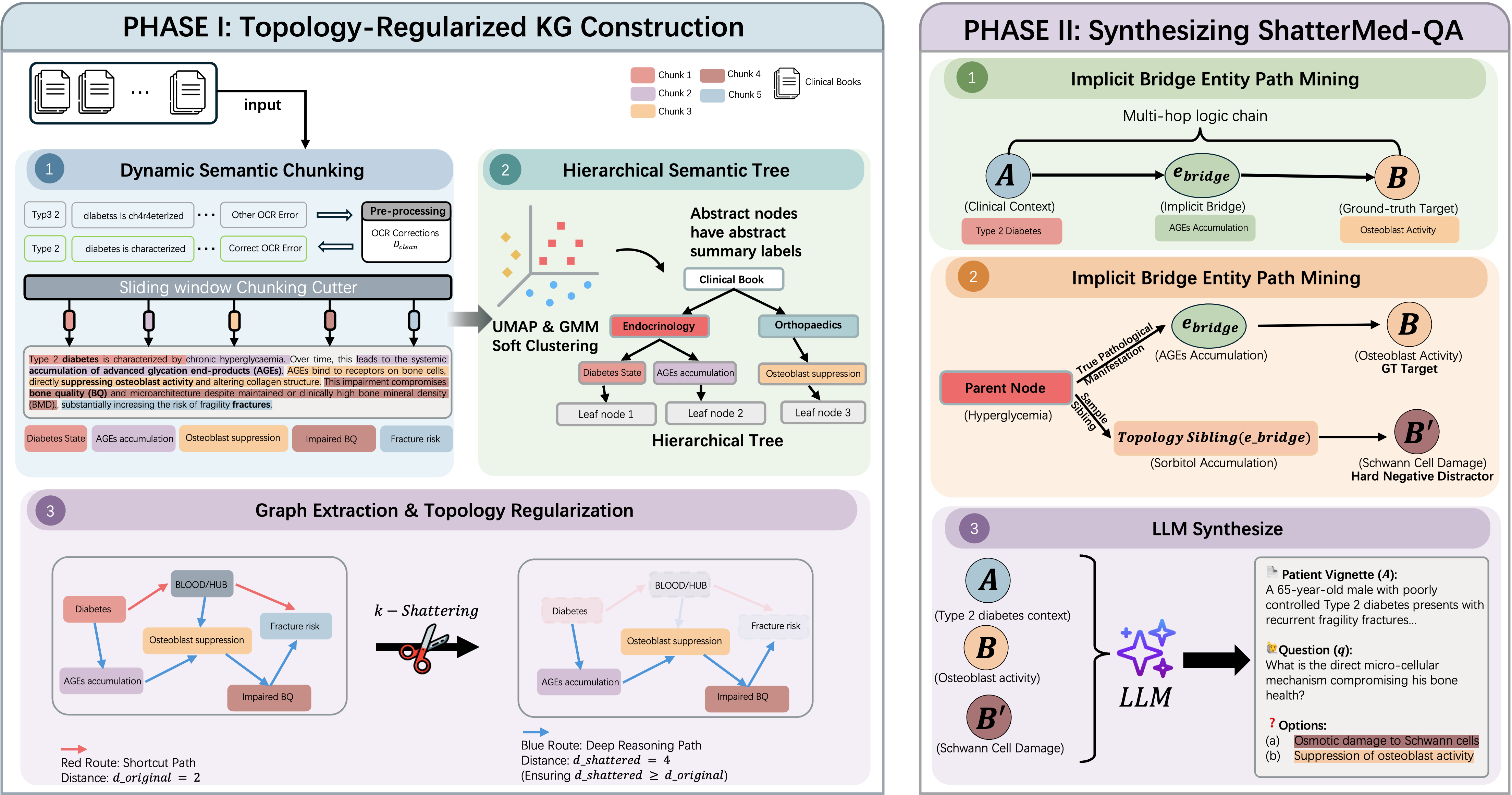}
    \caption{The end-to-end methodological pipeline of ShatterMed-QA.}
    \label{fig:pipeline}
\end{figure*}

\section{Methodology}

Existing medical datasets primarily evaluate shallow factual retrieval, failing to assess Large Language Models (LLMs) in complex diagnostic scenarios. To address this, we propose an end-to-end framework, visually outlined in \textbf{Figure \ref{fig:pipeline}} and formally orchestrated by \textbf{Algorithm \ref{alg:framework}}, to construct a topology-regularized medical Knowledge Graph (KG) and synthesize a multi-hop reasoning benchmark.

\begin{algorithm}[tb]
\caption{End-to-End Medical KG \& Benchmark Construction}
\label{alg:framework}
\textbf{Input}: Raw medical corpus $\mathcal{D}$, LLM \quad \textbf{Output}: Regularized KG $\mathcal{G}_{shatt}$, Benchmark $\mathcal{Q}$
\begin{algorithmic}[1]
\STATE \textbf{Phase I: Topology-Regularized KG Construction}
\STATE $\mathcal{D}_{clean} \leftarrow \text{CleanOCR}(\mathcal{D})$
\STATE $\mathcal{N}^{(0)} \leftarrow \text{SemanticChunk}(\mathcal{D}_{clean}, \text{threshold}=P_{95})$ \COMMENT{Preserve logic via cosine dist.}
\STATE $\mathcal{T} \leftarrow \text{HierarchicalTree}(\mathcal{N}^{(0)}, \text{GMM}, \text{BIC})$ \COMMENT{Soft clustering}
\STATE $\mathcal{G} \leftarrow \text{ExtractAndAlignTriplets}(\mathcal{T})$
\STATE $\mathcal{G}_{shatt} \leftarrow \text{PruneHubs}(\mathcal{G}, k, \text{Stoplist})$ \COMMENT{$k$-Shattering regularizes shortcuts}
\STATE \textbf{Phase II: QA Synthesis}
\STATE $\mathcal{Q} \leftarrow \emptyset$
\FOR{\textbf{each} constrained path $(A \rightarrow e_{bridge} \rightarrow B) \in \mathcal{G}_{shatt}$}
    \STATE $B' \leftarrow \text{SampleSiblingDistractor}(e_{bridge})$ \COMMENT{Hard Negative Sampling}
    \STATE $\mathcal{Q} \leftarrow \mathcal{Q} \cup \{\text{SynthesizeVignette}(A, B, B') \mid e_{bridge} \notin q\}$
\ENDFOR
\STATE \textbf{return} $\mathcal{G}_{shatt}, \mathcal{Q}$
\end{algorithmic}
\end{algorithm}

\subsection{Preliminary: The Pitfall of Shortcut Learning}
\label{sec:preliminary}

Let a medical Knowledge Graph be defined as $\mathcal{G} = (\mathcal{V}, \mathcal{E})$. Existing benchmarks mostly evaluate single-hop reasoning ($e_s \rightarrow e_t$), allowing models to guess answers via superficial pattern matching. In contrast, clinical diagnostics require multi-hop reasoning across a relational chain $P = (e_s \rightarrow \dots \rightarrow e_t)$.

We specifically target 2-hop chains connecting a source clinical context $A \in \mathcal{V}$ to a diagnostic target $B \in \mathcal{V}$. This connection relies on an intermediate node acting as a vital pathological mechanism, defined as the \textit{implicit bridge entity} ($e_{bridge} \in \mathcal{V}$), creating the target cascade $A \rightarrow e_{bridge} \rightarrow B$.

A critical vulnerability arises because natural KGs are dominated by highly connected generic hub nodes, defined as the subset $\mathcal{H} \subset \mathcal{V}$ (e.g., ''Blood'', ''Inflammation''). When evaluated on the connection between $A$ and $B$, LLMs often exploit these hubs to construct spurious logical shortcuts, bypassing the true micro-pathology $e_{bridge}$. This collapses complex reasoning into superficial retrieval via a generic hub $h$, where $A \rightarrow h \rightarrow B \quad (h \in \mathcal{H})$. Our framework mathematically regularizes the network topology to physically sever these shortcuts.

\subsection{Phase I: Topology-Regularized KG Construction}

As detailed in Phase I of Algorithm \ref{alg:framework} and Figure \ref{fig:pipeline}, we reconstruct a macroscopic clinical context from raw text while strictly eliminating topological shortcuts. 

\textbf{Preserving Semantic Integrity (Lines 2-4):} Conventional fixed-length token chunking blindly divides text based on arbitrary token limits. This fundamentally risks severing contiguous clinical causal chains, such as separating a disease's etiology from its resulting symptoms, before the knowledge is even extracted. To ensure that complete pathological mechanisms are ingested as cohesive units, we transition from length-based to semantic-based boundary detection. By enforcing breakpoints strictly where the cosine distance between consecutive sentence embeddings exceeds a dynamic 95th-percentile threshold ($\tau$), we encapsulate entire clinical cascades within individual chunks (e.g., preserving the complete context of ''\textit{AGEs accumulation}''). Furthermore, because isolated chunks lack macroscopic context, we draw inspiration from recent hierarchical retrieval paradigms~\cite{sarthi2024raptor,gupta2025llmguidedhierarchicalretrieval} to aggregate them into a hierarchical semantic tree ($\mathcal{T}$). To mitigate the curse of dimensionality, chunk embeddings are first projected via UMAP before applying Gaussian Mixture Models (GMM) optimized by the Bayesian Information Criterion (BIC)~\cite{umap}. This probabilistic soft clustering naturally accommodates the reality that a single medical concept inherently belongs to multiple overlapping disciplines.

\textbf{$k$-Shattering Regularization (Lines 5-6):} To physically prune the anomalous hub set $\mathcal{H}_k$, we apply a deterministic pre-construction pruning strategy. Specifically, $k$ acts as a global occurrence frequency threshold ($k=50$ in our implementation); entities exceeding this frequency across the corpus, alongside a meticulously curated clinical stop-entity list (e.g., generic terms like "patient" or "treatment"), are aggressively discarded prior to edge formation. By eliminating paths intersecting these ubiquitous hubs ($\mathcal{G}_{shatt}$), the monotonicity of the minimum function guarantees the reconstructed shortest path strictly increases or remains equal to the original distance:
$$ d_{shattered}(u,v) \ge d_{original}(u,v) $$
This structural constraint (detailed proof in \textbf{Supplementary \ref{supp:shattering}}) helps reduce trivial shortcut paths. For example, as illustrated in the bottom-left of \textbf{Figure \ref{fig:pipeline}}, pruning the generic \textit{Blood} hub removes the original short path ($d_{original}=2$). The retained path therefore traverses a more specific micro-pathological cascade ($d_{shattered}=4$): \textit{Type 2 Diabetes} $\rightarrow$ \textit{AGEs accumulation} $\rightarrow$ \textit{Osteoblast suppression} $\rightarrow$ \textit{Impaired Bone Quality} $\rightarrow$ \textit{Fracture risk}.

\subsection{Phase II: Synthesizing Constrained Diagnostics}

With the regularized $\mathcal{G}_{shatt}$ established, Phase II of Algorithm \ref{alg:framework} synthesizes the benchmark ($\mathcal{Q}$) by mining the constrained paths through a cyclical pipeline (depicted on the right side of \textbf{Figure \ref{fig:pipeline}}).

For each 2-hop chain ($A \rightarrow e_{bridge} \rightarrow B$), an LLM drafts a patient vignette based on clinical context $A$ and inquires about target $B$. To rigorously evaluate exclusionary reasoning, two explicit constraints are enforced (Lines 10-11): (1) \textbf{Implicit Masking:} The exact term for $e_{bridge}$ is strictly masked, compelling internal deduction ($e_{bridge} \notin q$). (2) \textbf{Topology-Driven Distractors:} To prevent models from succeeding via superficial elimination, we sample a sibling node within $e_{bridge}$'s pathological hierarchy. As demonstrated in \textbf{Figure \ref{fig:pipeline} (Phase II)}, rather than relying on the masked bridge \textit{AGEs Accumulation}, we query its topological sibling, \textit{Sorbitol Accumulation}. The downstream target of this sibling is then extracted to serve as the hard negative distractor ($B'$: \textit{Schwann Cell Damage}).

Because both $B$ (Osteoblast Activity) and $B'$ (Schwann Cell Damage) are genuine, biologically plausible complications of Diabetes, models relying on ungrounded statistical associations will fail. They must perform deep, exclusionary micro-pathological reasoning to successfully navigate ShatterMed-QA.

\section{Dataset Statistics and Task Distribution}
\label{sec:dataset_stats}

\begin{figure}[htbp]
    \centering
    \begin{subfigure}[b]{0.47\textwidth}
        \centering
        \includegraphics[width=\textwidth]{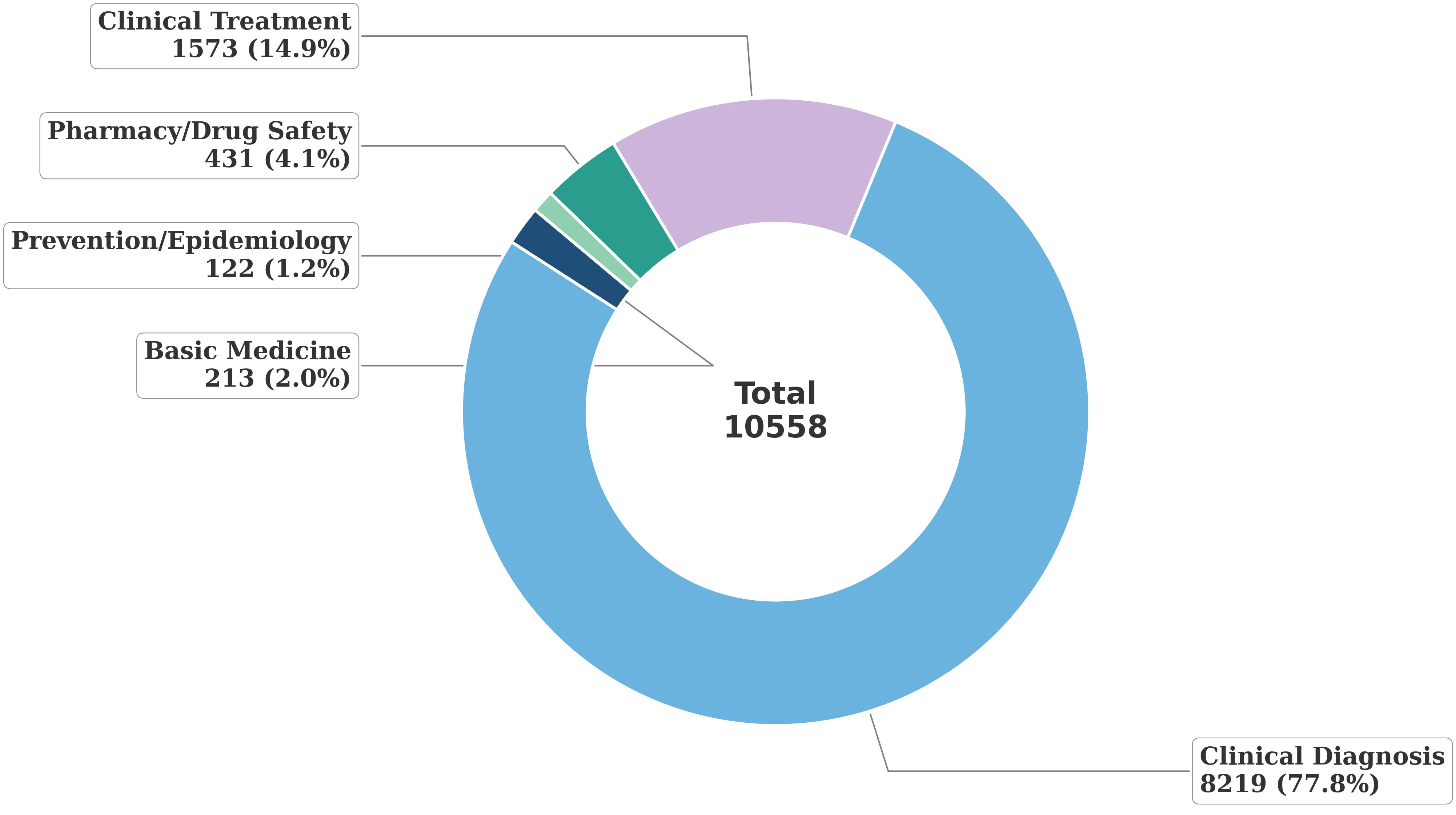}
        \caption{Overall distribution of the 10,558 tasks.}
        \label{fig:tasks_pie}
    \end{subfigure}
    \hfill
    \begin{subfigure}[b]{0.52\textwidth}
        \centering
        \includegraphics[width=\textwidth]{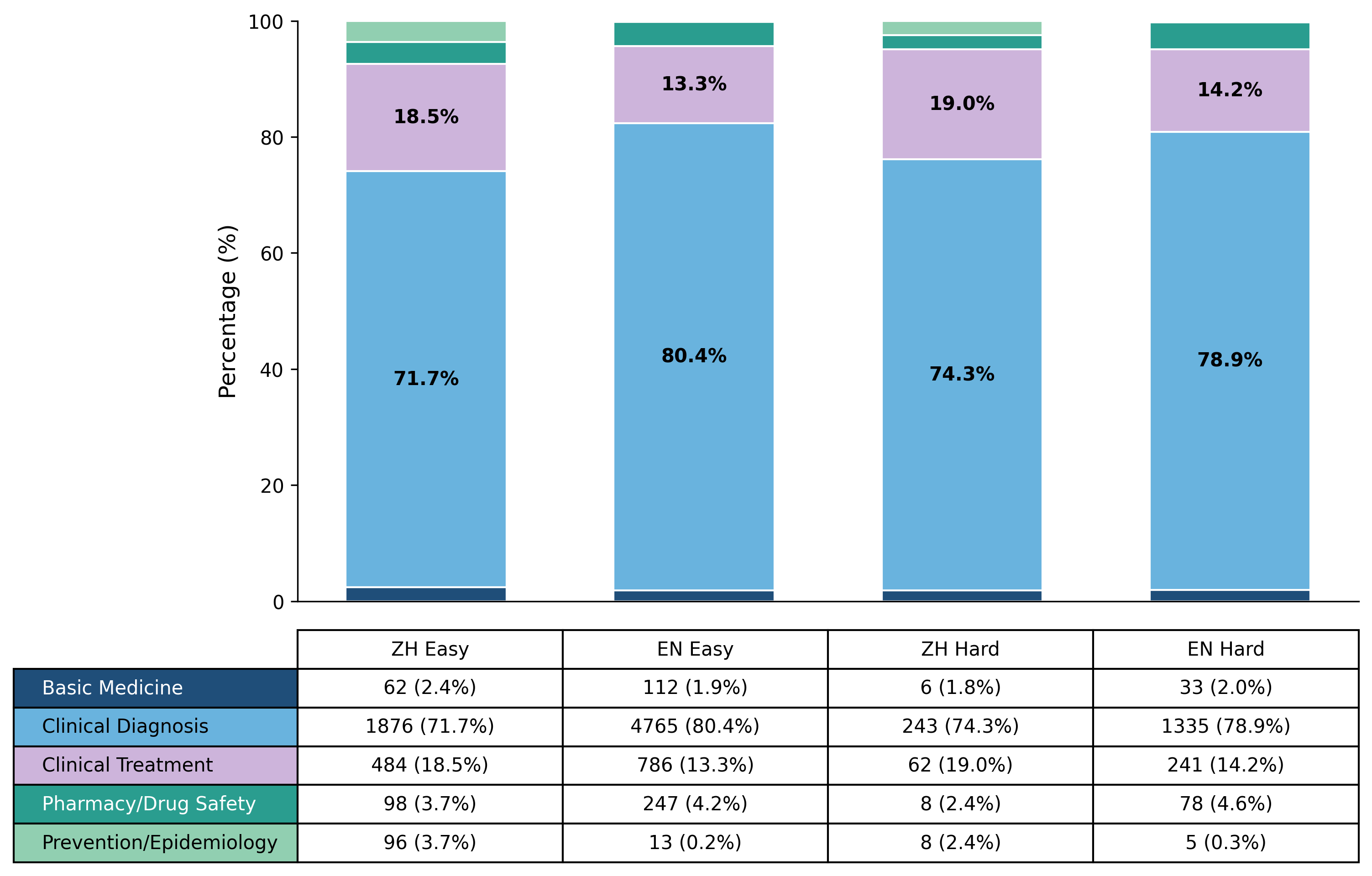}
        \caption{Fine-grained distribution across data splits.}
        \label{fig:tasks_bar}
    \end{subfigure}
    \caption{Distribution of clinical tasks within the ShatterMed-QA benchmark. (a) illustrates the macroscopic task proportions heavily anchored in clinical diagnosis, while (b) demonstrates the structural consistency across different languages (ZH/EN) and difficulty tiers (Easy/Hard).}
    \label{fig:task_distribution}
\end{figure}

To comprehensively evaluate the multi-hop diagnostic reasoning capabilities of Large Language Models (LLMs), ShatterMed-QA comprises a total of 10,558 meticulously synthesized clinical questions. The dataset is categorized into five primary clinical tasks, reflecting a diverse spectrum of real-world medical scenarios, as illustrated in Figure \ref{fig:task_distribution}.

As shown in Figure \ref{fig:task_distribution}(a), the benchmark is intentionally dominated by Clinical Diagnosis tasks, which account for the vast majority of the dataset with 8,219 questions (77.8\%). This deliberate concentration aligns perfectly with our core objective: real-world medical AI systems fail most catastrophically during complex diagnostic deduction, not during simple factual recall. By heavily anchoring ShatterMed-QA in diagnosis, we explicitly formulate a stress-test for a model's ability to navigate deep, multi-hop pathological reasoning and resist shortcut learning. Clinical Treatment represents the second largest category (1,573 questions, 14.9\%), evaluating the model's ability to deduce appropriate therapeutic interventions based on derived diagnoses. The remaining minor tasks include Pharmacy/Drug Safety (4.1\%), Basic Medicine (2.0\%), and Prevention/Epidemiology (1.2\%), ensuring a well-rounded but heavily diagnostic-focused medical knowledge assessment.

Furthermore, to ensure evaluation robustness, ShatterMed-QA is bifurcated into cross-lingual (English and Chinese) and difficulty-based (Easy and Hard) splits. Figure \ref{fig:task_distribution}(b) details the fine-grained task distribution across these four distinct subsets. Notably, the structural proportions remain remarkably consistent. For instance, Clinical Diagnosis consistently dominates across the ZH Easy (71.7\%), EN Easy (80.4\%), ZH Hard (74.3\%), and EN Hard (78.9\%) splits. This consistent proportional distribution mathematically guarantees that model performance variations observed between splits are driven by intrinsic reasoning difficulty (e.g., the masking of implicit bridge entities) rather than skewed task distributions.

\begin{table}[tb]
\caption{Overall statistics of the synthesized clinical QA dataset. Lengths are measured in words for English (EN) and characters for Chinese (ZH). Lexical overlaps quantify the exact matching between generated questions and source knowledge. Scores for Clarity, Validity, and Difficulty represent the mean results from a consensus-based ensemble evaluation.}
\label{tab:dataset_stats}
\centering
\begin{tabular}{lcccc c}
\toprule
\multirow{2}{*}{\textbf{Metric}} & \multicolumn{2}{c}{\textbf{English (EN)}} & \multicolumn{2}{c}{\textbf{Chinese (ZH)}} & \multirow{2}{*}{\textbf{Overall}} \\
\cmidrule(lr){2-3} \cmidrule(lr){4-5}
 & \textbf{Easy} & \textbf{Hard} & \textbf{Easy} & \textbf{Hard} & \\
\midrule
Total QA Pairs & 5,923 & 1,692 & 2,616 & 327 & 10,558 \\
Avg. Question Length & 155.9 & 199.9 & 77.5 & 107.2 & - \\
Avg. Explanation Length & 805.3 & 869.8 & 629.7 & 593.9 & - \\
\midrule
Avg. Distractor Similarity (Cosine) & 0.579 & 0.606 & 0.629 & 0.658 & 0.598 \\
Lexical Overlap vs. Content A (ROUGE-1) & 0.096 & 0.104 & 0.081 & 0.076 & 0.093 \\
Lexical Overlap vs. Content B (ROUGE-1) & 0.103 & 0.113 & 0.088 & 0.082 & 0.100 \\
Lexical Overlap vs. Evidence (BLEU-1) & 0.704 & 0.681 & 0.412 & 0.311 & 0.609 \\
\midrule
Ensemble-Judged Clarity (1-5) & 4.72 & 4.60 & 4.71 & 4.52 & 4.70 \\
Ensemble-Judged Validity (1-5) & 4.89 & 4.83 & 4.84 & 4.73 & 4.87 \\
Ensemble-Judged Difficulty (1-5) & 3.11 & 3.11 & 3.15 & 3.18 & 3.11 \\
\bottomrule
\end{tabular}
\end{table}

\subsection{Dataset Quality and Lexical Analysis}
\label{subsec:dataset_quality}
To empirically validate the rigor and quality of ShatterMed-QA, we conducted a comprehensive statistical analysis as summarized in Table \ref{tab:dataset_stats}. A fundamental concern in traditional clinical QA datasets is information leakage caused by high lexical overlap between questions and source material. Our analysis demonstrates that the ROUGE-1 overlap between synthesized questions and source knowledge (Content A and B) is exceptionally low, averaging 0.093 and 0.100, respectively. This mathematically validates our implicit bridge entity masking strategy ($e_{bridge} \notin q$), ensuring that models cannot succeed via superficial string matching but must instead comprehend the underlying clinical vignette to infer the hidden pathology. Furthermore, the effectiveness of our hard negatives is evidenced by an average distractor similarity of approximately 0.598. This high cosine similarity confirms that topology-driven sibling sampling successfully retrieves biologically plausible, semantically adjacent distractors, thereby neutralizing simple elimination strategies.To ensure clinical soundness and high-quality taxonomy, we employed a consensus-based automated quality assurance pipeline utilizing an ensemble of three frontier LLMs (GPT-5/4.1-mini, Grok-4-1) acting as independent expert adjudicators. The final quality scores reflect the arithmetic mean of these evaluations across three critical dimensions. The benchmark achieved a robust clarity score of 4.70, indicating that the questions are unambiguous and well-formulated, while a validity score of 4.87 confirms high clinical reliability and medically sound rationales. Finally, an average difficulty score of 3.11 indicates a challenging complexity level that is well-suited for evaluating the deep reasoning capabilities of frontier LLMs.

\subsection{Expert Human Validation}
\label{sec:human_validation}

To evaluate the clinical validity of ShatterMed-QA, three board-certified physicians from Grade 3A hospitals in China conducted a blind review of 700 sampled questions. Our detailed analysis focused on the Chinese (ZH) Hard split, consisting of 327 multi-hop diagnostic vignettes. 

Out of the 327 ZH-Hard questions, 21 (6.4\%) were identified as incorrect. In addition, 9 cases showed temporal conflicts, where the generated reasoning aligned with standard textbook logic but differed from updated clinical guidelines. The physicians also identified 33 questions as clinically ambiguous. In these instances, the ground truth (GT) represents the standard textbook answer, but alternative options may be justifiable depending on patient-specific contexts. For example, as shown in the anesthesia case study, clinical treatment planning can be individualized and may lack a single standard answer. Excluding these 63 flagged cases, the remaining 264 questions (80.7\%) in the ZH-Hard split were verified by the physicians as clinically valid and fully usable for evaluation. Therefore, while the generation pipeline maintains logical consistency based on textbooks, it does not capture all clinical variability.

\subsubsection{Limitations.} There are certain limitations to this human validation process. First, because clinical medicine is highly specialized, evaluating cross-disciplinary reasoning paths occasionally required the reviewers to consult external medical literature to verify the ground truth. Second, as the clinical experts practice in China, their assessments may reflect regional clinical practices and guidelines, potentially introducing geographic bias when evaluating questions based on overseas practices. Despite these constraints, the manual review indicates that ShatterMed-QA serves as a valid, textbook-grounded benchmark for medical reasoning.

\begin{table}[!t]
\caption{Comparison of ShatterMed-QA with existing datasets. Automated proxy metrics (Clarity, Validity, Difficulty) are derived via a zero-shot LLM adjudicator. \textit{*Note: Although the scores are from three different models, these scores serve as baseline quality filters and may exhibit generative self-preference bias.}}
\label{tab:comprehensive_comparison}
\centering
\footnotesize 
\begin{threeparttable}
\begin{tabular}{@{}llccccc@{}}
\toprule
\textbf{Type} & \textbf{Dataset} & \textbf{Traceability} & \textbf{Size} & \textbf{Clarity} & \textbf{Validity} & \textbf{Diff.} \\ \midrule

\multirow{4}{*}{\shortstack[l]{\textbf{Real-World}\\\textbf{Exams}}} 
 & MedQA~\cite{jin2021disease}\tnote{\dag} & Chapter & 6,110 & 3.92 & 3.87 & 2.96 \\
 & MMedBench~\cite{qiu2024towards}\tnote{\ddag} & Chapter & 8,518 & 4.03 & 4.04 & 2.79 \\
 & MultiMedQA~\cite{singhal2025toward}\tnote{\S} & Chapter & 1,242 & 3.93 & 3.94 & 2.64 \\
 & AfrimedQA~\cite{olatunji2024afrimed} & Chapter & 3,723 & 3.80 & 3.72 & 2.83 \\ \midrule

\textbf{Research} 
 & PubMedQA~\cite{jin2019pubmedqa} & Abstract & 1,000 & 2.92 & 2.43 & 2.46 \\ \midrule

\multirow{3}{*}{\textbf{AI-Assisted}} 
 & CRAFT-MedQA~\cite{ziegler2025craft} & LLM Rationale & 5,000 & 3.71 & 3.52 & 2.55 \\
 & MedReason~\cite{wu2025medreason} & Graph & 25,484 & 3.79 & 3.81 & 3.04 \\
 & MedQA-Evol~\cite{zhang2024ultramedical} & LLM Rationale & 51,809 & 4.17 & 4.32 & 2.94 \\ \midrule

\textbf{Ours} 
 & \textbf{ShatterMed-QA} & \textbf{Sentence \& Page} & \textbf{10,558} & \textbf{4.70} & \textbf{4.87} & \textbf{3.11} \\ \bottomrule
\end{tabular}
\begin{tablenotes}\footnotesize
\item \textit{Notes:} \textsuperscript{\dag,\ddag,\S} Weighted averages for regional exams (\dag), six languages (\ddag), and MMLU domains (\S). Sizes reflect downloaded configurations (breakdown in Supp. \ref{supp:dataset_breakdown}).
\end{tablenotes}
\end{threeparttable}
\end{table}

\subsection{Comparison with Existing Benchmarks}
\label{sec:benchmark_comparison}

To evaluate ShatterMed-QA, we applied a zero-shot LLM-as-a-judge prompt across established datasets (Table \ref{tab:comprehensive_comparison}). While confirming high structural quality, we acknowledge the inherent self-preference bias of LLMs when evaluating AI-generated text~\cite{gu2024survey}; therefore, these scores serve strictly as a baseline quality filter. Unlike unconstrained datasets (e.g., CRAFT-MedQA's 3.52 validity), our topological constraints yield a peak automated validity of 4.87. Crucially, definitive clinical validity is anchored by our blind human evaluation (Section 4.2), which verified an 80.7\% clinical usability rate. Furthermore, while real-world exams necessarily include simple, single-hop questions resulting in lower difficulty scores (e.g., MedQA at 2.96), ShatterMed-QA achieves a higher automated difficulty of 3.11, validating that $k$-Shattering and implicit masking effectively sever trivial shortcuts. Finally, unlike human-curated datasets providing coarse chapter-level evidence, ShatterMed-QA anchors every synthesized reasoning chain to exact sentence-level text, successfully shifting evaluation from black-box accuracy to verifiable reasoning path analysis.

\section{Experiments and Evaluation Metrics}
\label{sec:experiments}
To assess intrinsic clinical generalization without task-specific fine-tuning, we evaluated 21 LLMs across three categories, including Proprietary Frontier, Open-Source General Foundation, and Domain-Specific Medical, using PyTorch on NVIDIA L40 GPUs. Utilizing Hugging Face checkpoints for local inference, our core protocol establishes a performance baseline via strict zero-shot multiple-choice accuracy paired with novel behavioral metrics to diagnose specific reasoning bottlenecks , with full model and configuration details deferred to Supplementary \ref{supp:exp_details}. Furthermore, the effect of our $k$-Shattering is examined via structural ablation: as $k$ is reduced from $\infty$ to 50, the average shortest path (ASP) in the ZH-KG increases from 3.549 to 3.603, suggesting that generic shortcut entities are partially suppressed and that retained reasoning paths become modestly less direct, with detailed graph statistics provided in Supplementary \ref{supp:ablation_stats}.

\subsection{Behavioral Metrics for Implicit Reasoning}
\label{sec:behavioral_metrics}

While zero-shot accuracy provides a macroscopic performance baseline, traditional text-overlap metrics (e.g., ROUGE, BLEU) frequently fail to capture logical validity in medical diagnostics. Furthermore, LLM-as-a-judge approaches introduce severe self-preference bias, and many models suffer from mode collapse, failing to generate explicit Chain-of-Thought (CoT) reasoning when constrained~\cite{gu2024survey}. To rigorously evaluate whether models are executing genuine multi-hop reasoning rather than relying on statistical heuristics, we introduce two targeted, black-box behavioral metrics based on discrete categorical counting.

\subsubsection{Hard Negative Error Rate (HNE)} 
This metric quantifies a model's susceptibility to topology-driven shortcut learning. When a model fails a question, HNE measures the exact proportion of times it is deceived by the biologically plausible, sibling distractor ($B'$) rather than other random incorrect options. Let $E_{wrong}$ be the set of all incorrect predictions for a given model. HNE is defined as:
$$HNE = \frac{\sum_{i \in E_{wrong}} \mathbb{I}(\text{Prediction}_i = B'_i)}{|E_{wrong}|}$$
In a standard four-option setup, uniform random guessing among the three incorrect options yields an expected baseline rate of $\approx 33.3\%$. An HNE significantly exceeding this threshold provides mathematical evidence that the model is bypassing the implicit bridge entity and relying on single-hop superficial associations.

\subsubsection{Reasoning Recovery Rate ($R^3$)} 
To decouple fundamental reasoning deficits from mere parametric knowledge gaps, we utilize $R^3$. It calculates the percentage of zero-shot failures that are successfully corrected when the masked implicit bridge entity ($e_{bridge}$) is explicitly supplied via Retrieval-Augmented Generation (RAG). To simulate a realistic and challenging retrieval environment, the golden source paragraph containing $e_{bridge}$ is randomly embedded among $k=5$ retrieved background documents. 
$$R^3 = \frac{|\text{Incorrect in Zero-Shot} \cap \text{Correct in RAG}|}{|\text{Incorrect in Zero-Shot}|}$$
A high $R^3$ validates the structural fidelity of the dataset, indicating the multi-hop logic is sound and the model can successfully perform context synthesis once the topological knowledge gap is bridged. Conversely, a low $R^3$ exposes a fundamental collapse in a model's reasoning engine.

\begin{table}[!t]
\caption{Zero-shot (closed-book) accuracy (\%) of evaluated models on ShatterMed-QA.}
\label{tab:direct_results}
\centering
\resizebox{\textwidth}{!}{
\begin{tabular}{@{}llcccc@{}}
\toprule
\multirow{2}{*}{\textbf{Model}} & \multirow{2}{*}{\textbf{Size}} & \multicolumn{2}{c}{\textbf{English (EN)}} & \multicolumn{2}{c}{\textbf{Chinese (ZH)}} \\ \cmidrule(lr){3-4} \cmidrule(l){5-6} 
 & & \textbf{Easy} & \textbf{Hard} & \textbf{Easy} & \textbf{Hard} \\ \midrule

\textbf{Proprietary Frontier Models} & & & & & \\
GPT-5-mini~\cite{gpt5} & -- & 97.45 & 96.81 & 97.82 & 96.33 \\
GPT-5-nano~\cite{gpt5} & -- & 91.59 & 89.83 & 91.13 & 86.54 \\
GPT-4.1-mini~\cite{gpt41} & -- & \textbf{98.31} & 97.28 & 98.17 & \textbf{96.94} \\
GPT-4.1-nano~\cite{gpt41} & -- & 96.25 & 93.85 & 92.35 & 84.71 \\
Grok-4.1-fast-reasoning~\cite{grok41} & -- & 98.08 & 97.58 & 98.24 & 96.64 \\
Grok-4.1-fast-non-reasoning~\cite{grok41} & -- & 74.35 & 96.10 & 97.63 & 96.33 \\
Grok-4-fast-reasoning~\cite{grok4} & -- & 98.18 & \textbf{97.87} & \textbf{98.59} & 96.64 \\
Grok-4-fast-non-reasoning~\cite{grok4} & -- & 96.96 & 96.04 & 97.82 & 96.02 \\ \midrule

\textbf{Open-Weights (General Foundation)} & & & & & \\
Qwen3-14B-Base~\cite{yang2025qwen3} & 14B & \textbf{90.04} & \textbf{86.47} & 95.11 & \textbf{92.35} \\
Falcon3-10B-Base~\cite{site_falcon3_10b_base} & 10B & 85.80 & 83.70 & 76.86 & 63.61 \\
Yi-1.5-9B~\cite{young2024yi} & 9B & 79.23 & 69.17 & 88.09 & 80.73 \\
Gemma-2-9b~\cite{team2024gemma} & 9B & 77.17 & 55.82 & 77.79 & 79.51 \\
Llama-3.1-8B~\cite{grattafiori2024llama} & 8B & 79.40 & 72.30 & 72.69 & 67.89 \\
InternLM3-8B-Instruct~\cite{site_internlm3_8b_instruct} & 8B & 80.20 & 78.44 & 91.66 & 89.30 \\
c4ai-command-r7b-12-2024~\cite{site_command_r7b_122024} & 7B & 77.70 & 71.59 & 79.03 & 74.31 \\
Granite-3.3-8b-base~\cite{site_granite33_8b_base} & 8B & 68.12 & 65.27 & \textbf{96.39} & 70.95 \\ \midrule

\textbf{Domain-Specific (Medical)} & & & & & \\
Llama3-Med42-8B~\cite{christophe2024med42} & 8B & \textbf{82.64} & \textbf{80.45} & 85.81 & \textbf{78.90} \\
MedGemma-1.5-4b-it~\cite{sellergren2025medgemma} & 4B & 81.85 & 76.26 & 77.55 & 74.62 \\
Meditron-7B~\cite{bosselut2024meditron} & 7B & 80.60 & 64.74 & \textbf{92.34} & 73.09 \\
Llama3-OpenBioLLM-8B~\cite{site_openbiollm_llama3_8b} & 8B & 73.37 & 63.14 & 62.55 & 55.05 \\
BioMistral-7B~\cite{labrak2024biomistral} & 7B & 60.59 & 54.64 & 52.17 & 44.04 \\ \bottomrule
\end{tabular}
}
\end{table}

\begin{table}[tb]
\caption{Aggregated Behavioral Metrics: Hard Negative Error Rate (HNE) and Reasoning Recovery Rate ($R^3$). Models are sorted by their susceptibility to topological shortcuts (HNE). A random guessing baseline for HNE is $\approx 33.3\%$.}
\label{tab:hne_r3_results}
\centering
\resizebox{\textwidth}{!}{
\begin{tabular}{@{}lccc@{}}
\toprule
\textbf{Model} & \textbf{Total Zero-Shot Errors} & \textbf{HNE Rate ($\uparrow$ worse)} & \textbf{$R^3$ Rate ($\uparrow$ better)} \\ \midrule

\textbf{Proprietary Frontier} & & & \\
gpt-5-mini & 66 & \textbf{53.03\%} & 66.67\% \\
gpt-5-nano & 216 & 43.52\% & 51.39\% \\
gpt-4.1-mini & 56 & 41.07\% & \textbf{69.64\%} \\
grok-4-fast-reasoning & 47 & 38.30\% & 61.70\% \\
grok-4-fast-non-reasoning & 80 & 37.50\% & 67.50\% \\
grok-4-1-fast-non-reasoning & 78 & 37.18\% & 61.54\% \\
grok-4-1-fast-reasoning & 52 & 36.54\% & 61.54\% \\
gpt-4.1-nano & 154 & 31.82\% & 48.70\% \\ \midrule

\textbf{Open-Weights (General)} & & & \\
Llama-3.1-8B & 574 & \textbf{45.47\%} & 61.67\% \\
Qwen3-14B-Base & 254 & 41.73\% & 46.85\% \\
Yi-1.5-9B & 585 & 40.85\% & 46.50\% \\
internlm3-8b-instruct & 400 & 37.75\% & 38.00\% \\
Falcon3-10B-Base & 395 & 37.72\% & 31.90\% \\
gemma-2-9b & 815 & 37.42\% & 63.44\% \\
granite-3.3-8b-base & 683 & 36.90\% & 29.28\% \\
c4ai-command-r7b-12-2024 & 565 & 34.69\% & 37.88\% \\ \midrule

\textbf{Domain-Specific (Medical)} & & & \\
medgemma-1.5-4b-it & 485 & \textbf{39.79\%} & 37.94\% \\
Llama3-Med42-8B & 400 & 38.75\% & 34.00\% \\
meditron-7b & 685 & 36.50\% & \textbf{7.30\%} \\
Llama3-OpenBioLLM-8B & 771 & 26.07\% & 48.12\% \\
BioMistral-7B & 951 & 25.55\% & \textbf{60.78\%} \\ \bottomrule
\end{tabular}
}
\end{table}

\section{Results and Analysis}
\subsection{Zero-Shot Baseline Performance}
Table \ref{tab:direct_results} presents the performance of benchmarked models under a closed-book, direct inference setting. We observe a systemic performance degradation as models transition from the Easy to the Hard split, particularly among open-weights architectures. For instance, \textit{Gemma-2-9b} exhibits a dramatic 21.35\% drop in English accuracy (from 77.17\% to 55.82\%), and \textit{BioMistral-7B} falls to 44.04\% on the Chinese Hard split. This consistent gap empirically validates the efficacy of our $k$-Shattering pipeline in eliminating logical shortcuts. Notably, while proprietary frontier models maintain near-perfect stability across splits, many domain-specific medical LLMs are outperformed by general foundation models like \textit{Qwen3-14B-Base}, which achieves 92.35\% on ZH-Hard. This suggests that current medical fine-tuning strategies often prioritize factual recall over the deep, multi-hop diagnostic reasoning required by ShatterMed-QA.

\subsection{The Allure of the Shortcut: HNE Analysis}
\label{sec:hne_analysis}

Table \ref{tab:hne_r3_results} summarizes the behavioral metrics for the Hard split, revealing a systemic vulnerability to shortcut learning. In a four-option setup, a random-guessing baseline for the three incorrect options is 33.3\%. However, frontier models actively gravitate toward topological traps, with GPT-5-mini and Llama-3.1-8B recording HNE rates of 53.03\% and 45.47\%, respectively. This significant deviation confirms that models rely on single-hop, generic correlations rather than failing randomly. Such results corroborate that current models exploit generic knowledge hubs (e.g., linking "surgery" to "long duration") instead of deducing the authentic micro-pathological cascades required by ShatterMed-QA.

\subsection{Diagnosing Reasoning Deficits via $R^3$}
\label{sec:r3_analysis}

While zero-shot accuracy establishes a baseline, the Reasoning Recovery Rate ($R^3$) serves as a definitive probe to distinguish between knowledge absence and reasoning engine failure. As shown in Table \ref{tab:hne_r3_results}, providing the masked implicit evidence via RAG triggers substantial recoveries across most architectures. For example, BioMistral-7B initially failed 951 queries, yet successfully recovered $60.78\%$ of them when the topological gap was externally bridged. This widespread recovery empirically validates the structural integrity of the ShatterMed-QA vignettes; the multi-hop paths are logically sound and solvable, proving that initial failures stem from internal parametric gaps rather than flawed dataset synthesis.

Crucially, $R^3$ effectively exposes anomalous model architectures. Despite exhibiting a relatively standard error distribution (HNE of $36.50\%$), Meditron-7B yields a catastrophic $R^3$ of merely $7.30\%$. This indicates a severe limitation in context synthesis: even when explicitly provided with the missing pathophysiological links, the model is unable to execute the multi-hop deduction. This highlights a critical flaw in certain medical fine-tuning strategies, where models may overfit to static knowledge retrieval at the complete expense of dynamic logical reasoning. To concretely illustrate this shortcut learning phenomenon and the subsequent RAG-induced recovery, we provide a detailed clinical vignette analysis in Supplementary \ref{supp:case_study_details}.

\section{Conclusion}
\label{sec:conclusion}

We introduced ShatterMed-QA, a topology-regularized benchmark designed to rigorously evaluate multi-hop diagnostic reasoning in LLMs. To combat shortcut learning, our framework utilizes dynamic semantic chunking and a $k$-Shattering algorithm to physically prune generic hub nodes, forcing models to navigate authentic micro-pathological cascades. We further ensure a challenging, traceable baseline by employing implicit bridge entity masking and topology-driven hard negative sampling to generate biologically plausible distractors.

Evaluating 21 LLMs revealed systemic multi-hop reasoning deficits. Our novel behavioral metrics demonstrated that rather than guessing randomly, models actively gravitate toward single-hop topological shortcuts, with error rates on hard distractors exceeding 50\%. Crucially, a robust performance recovery under Retrieval-Augmented Generation (RAG), achieving recovery rates of nearly 70\%, validates our benchmark's structural fidelity, proving these failures stem from internal parametric knowledge gaps rather than flawed synthetic logic. Bolstered by expert human validation, ShatterMed-QA shifts medical AI evaluation from shallow recall to deep, exclusionary reasoning. Future work will integrate real-world clinical variability and leverage this topology-regularized framework to develop more robust medical fine-tuning strategies.

%
%

\clearpage
\appendix
\section*{Supplementary Material}

This supplementary material provides the detailed mathematical formulations, extended proofs, and concrete clinical examples that support the end-to-end framework presented in the main manuscript.

\section{Detailed Methodology}

\subsection{Detailed Formulation of the Hierarchical Semantic Tree}
\label{supp:gmm}

In Phase I of our methodology, constructing a hierarchical semantic summary tree requires projecting high-dimensional chunks into a computable space and clustering them into macroscopic medical themes. To address the \textit{curse of dimensionality}, where distance metrics lose discriminative power in ultra-high-dimensional spaces, we first apply Uniform Manifold Approximation and Projection (UMAP) to project the node embeddings onto a low-dimensional Riemannian manifold $\mathcal{Z} = \{\mathbf{z}_1, \dots, \mathbf{z}_N\}$.

Subsequently, we employ a Gaussian Mixture Model (GMM) for probabilistic soft clustering. Unlike the hard, non-overlapping assignments of K-Means, GMM naturally accommodates the overlapping nature of medical knowledge, allowing a text chunk to probabilistically belong to multiple medical topics simultaneously (e.g., a chunk might be 60\% relevant to cardiology and 40\% to nephrology). The log-likelihood function, parameterized by $\Theta = \{\pi_k, \mu_k, \Sigma_k\}_{k=1}^K$, is defined as:
\begin{equation}
\mathcal{L}(\Theta) = \sum_{i=1}^N \ln p(\mathbf{z}_i \mid \Theta) = \sum_{i=1}^N \ln \left( \sum_{k=1}^K \pi_k \mathcal{N}(\mathbf{z}_i \mid \mu_k, \Sigma_k) \right)
\end{equation}

Because the summation resides inside the logarithm, direct analytical differentiation is intractable. We optimize this objective using the Expectation-Maximization (EM) algorithm, which iteratively constructs and maximizes a lower bound of the likelihood function.

\textbf{E-step (Expectation):} We evaluate the posterior probability, or \textit{responsibility} $\gamma_{ik}$, that the $i$-th text chunk belongs to the $k$-th cluster. Intuitively, this calculates the empirical ''weight'' of each text in representing a specific medical theme:
\begin{equation}
\gamma_{ik} = \frac{\pi_k \mathcal{N}(\mathbf{z}_i \mid \mu_k, \Sigma_k)}{\sum_{j=1}^K \pi_j \mathcal{N}(\mathbf{z}_i \mid \mu_j, \Sigma_j)}
\end{equation}

\textbf{M-step (Maximization):} The algorithm leverages these responsibilities to recalibrate the parameters of each Gaussian distribution. We update the center location ($\mu_k$), the covariance matrix dictating the scope/variance ($\Sigma_k$), and the global mixing proportion ($\pi_k$) of each medical cluster to maximize the overall likelihood:
\begin{equation}
\mu_k = \frac{\sum_{i=1}^N \gamma_{ik} \mathbf{z}_i}{\sum_{i=1}^N \gamma_{ik}}
\end{equation}
\begin{equation}
\Sigma_k = \frac{\sum_{i=1}^N \gamma_{ik} (\mathbf{z}_i - \mu_k)(\mathbf{z}_i - \mu_k)^T}{\sum_{i=1}^N \gamma_{ik}}
\end{equation}
\begin{equation}
\pi_k = \frac{1}{N}\sum_{i=1}^N \gamma_{ik}
\end{equation}

\textbf{BIC Optimization:} Since the optimal number of medical topics $K$ is unknown \textit{a priori}, we must prevent the model from infinitely fragmenting clusters to artificially inflate the likelihood. We utilize the Bayesian Information Criterion (BIC) as a penalized objective function:
\begin{equation}
\text{BIC}(K) = -2 \ln(\hat{\mathcal{L}}) + K \ln(N)
\end{equation}
The algorithm dynamically searches for the optimal $K^*$ that minimizes $\text{BIC}(K)$. For each optimal cluster, an LLM generates a refined abstract summary, forming the parent node for the next hierarchical tier.

\subsection{Proof of Topology Regularization}
\label{supp:shattering}

To mitigate shortcut-prone reasoning paths, we apply topology-aware pruning via $k$-Shattering. Let $\mathcal{H}_k \subset \mathcal{V}$ denote the set of highly generic entities identified by the combined constraint of a predefined absolute frequency threshold $k$ and an expert-curated clinical stoplist (e.g., generic terms like "patient" or "inflammation"). In our implementation, these entities are excluded before edge formation so that they contribute less to downstream shortcut correlations. Let $\mathcal{P}_{u,v}$ be the set of all possible derivation paths between nodes $u$ and $v$ in the original graph $\mathcal{G}$. Their shortest path distance is defined as:
\begin{equation}
d_{original}(u,v) = \min \{ |p| \mid p \in \mathcal{P}_{u,v} \} \text{ }
\end{equation}

By pruning the set $\mathcal{H}_k$ during graph construction, shortcut paths that depend on these generic entities become unavailable in the retained path set. The new valid path set shrinks to $\mathcal{P}'_{u,v} = \{p \in \mathcal{P}_{u,v} \mid p \cap \mathcal{H}_k = \emptyset\}$. Consequently, the reconstructed shortest path distance $d_{shattered}(u,v)$ is defined as:

\begin{equation}
d_{shattered}(u,v) = \min \{ |p| \mid p \in \mathcal{P}'_{u,v} \} \text{ }
\end{equation}
Since the pruned path set is a strict subset of the original ($\mathcal{P}'_{u,v} \subseteq \mathcal{P}_{u,v}$), the monotonicity of the minimum function mathematically guarantees that the shortest distance between clinical context and diagnostic target can only increase or remain equal:
\begin{equation}
d_{shattered}(u,v) \ge d_{original}(u,v) \text{ }
\end{equation}

\textbf{Clinical Intuition and Case Study:} 
This inequality provides a structural guarantee that generic shortcut entities contribute less to the retained paths. By removing generalized ''highway'' terms, the reasoning process is encouraged to traverse more specific ''local roads'' (e.g., targeted enzymes or receptors). 

To illustrate, consider the causal link between \textit{Diabetes} ($u$) and \textit{Bone Fracture} ($v$). In the original graph, the model might exploit a trivial 2-hop shortcut via the ubiquitous hub \textit{Blood} ($h \in \mathcal{H}_k$): 
\begin{center}
\textit{Diabetes} $\rightarrow$ alters $\rightarrow$ \textit{Blood} $\rightarrow$ supplies $\rightarrow$ \textit{Bone Fracture} ($d_{original} = 2$). 
\end{center}
This explains nothing about the actual mechanism. After $k$-Shattering prunes \textit{Blood} and \textit{Inflammation}, the shortest path is forced to unravel the authentic micro-pathological cascade: 
\begin{center}
\textit{Diabetes} $\rightarrow$ accumulation of $\rightarrow$ \textit{Advanced Glycation End-products (AGEs)} $\rightarrow$ suppresses $\rightarrow$ \textit{Osteoblast activity} $\rightarrow$ compromises $\rightarrow$ \textit{Bone Mineral Density} $\rightarrow$ leads to $\rightarrow$ \textit{Fracture} ($d_{shattered} = 4$). 
\end{center}
This perfectly exemplifies $d_{shattered} \ge d_{original}$, guaranteeing that the regularized KG captures genuine clinical etiology rather than semantic coincidences.

\subsection{Entity Alignment Mechanism}
\label{supp:alignment}

To seamlessly integrate unstructured summary nodes from the LLM with the structured KG, we designed a dual-layer entity alignment system to resolve medical terminology polymorphism (synonyms, nested acronyms).

\textbf{Stage 1: Greedy MaxMatch.} Given an input character sequence $W = w_1 w_2 \dots w_m$ and a predefined graph vocabulary $\mathcal{V}$, the algorithm slides a window from left to right, greedily prioritizing the longest substring (maximum tokens) in $\mathcal{V}$ at each position, then advances to the end of the match. This strategy effectively prevents compound medical nouns from being erroneously tokenized into meaningless fragments.

\textbf{Stage 2: Fuzzy Edit Distance Soft Alignment.} We compute the Damerau-Levenshtein distance (allowing single-character insertions, deletions, substitutions, and adjacent transpositions) between a candidate string $s_1$ and a standard graph entity $s_2$:
\begin{equation}
DL(s_1, s_2) \le \theta(\text{len}(s_1))
\end{equation}
where $\theta$ is a dynamic tolerance threshold scaled by the source entity's length. Variants satisfying this threshold are forcibly merged into the standard entity, completely eliminating dangling nodes.

\subsection{Multi-Source Hard Negative Sampling Strategy}
\label{supp:hard_negative}

To prevent models from succeeding via superficial elimination or confirmation bias during Phase II, we introduce topology-driven hard negative sampling. When the path generation reaches the implicit bridge entity $e_{bridge}$ (e.g., \textit{Pneumonia}), the algorithm bifurcates to sample a sibling node within the same pathological hierarchy (e.g., branching to \textit{Tuberculosis}). The corresponding downstream node $B'$ (e.g., \textit{Isoniazid}) is then extracted as a distractor option. This high-barrier design guarantees that incorrect options are biologically plausible, forcing evaluated models to perform exhaustive micro-dimensional exclusionary reasoning.

\section{Extended Experimental Setup}
\label{supp:exp_details}

\subsection{Implementation and Hyperparameters}
All local open-weight and domain-specific models were deployed using their official checkpoints from Hugging Face. Computational experiments were executed on a localized cluster equipped with NVIDIA L40 GPUs utilizing the PyTorch framework.

\textbf{Phase I Graph Construction:} Dynamic semantic chunking was enforced using $\tau = P_{95}$. For $k$-Shattering, the global entity frequency threshold was set to $k=50$ as a practical operating point. Entities appearing $>50$ times across hierarchical nodes, along with those in our clinical stoplist (including generic terms like ''patient'', ''disease'', and chapter headers), were excluded from graph edge construction in order to suppress highly generic shortcut entities before downstream path mining. For the hierarchical soft clustering, the text embeddings were first projected. Subsequently, a Gaussian Mixture Model (GMM) was applied. The maximum number of permissible clusters was capped at 50, with the optimal number of topics ($K^*$) dynamically determined by minimizing the Bayesian Information Criterion (BIC).

\textbf{Retrieval-Augmented Generation (RAG) Pipeline:} The RAG knowledge integration process (detailed in Section \ref{sec:behavioral_metrics}) utilized a two-stage coarse-to-fine retrieval architecture. The initial coarse retrieval was performed using the \texttt{BAAI/bge-m3} embedding model, fetching an initial candidate pool of $K=50$ relevant chunks. These candidates were subsequently refined and re-scored by the \texttt{BAAI/bge-reranker-v2-m3} cross-encoder. The cross-encoder operated in FP16 precision with a batch size of 16, ultimately yielding the top $N=15$ most relevant documents to serve as the external evidence context for the LLMs.

\subsection{Topology Ablation Analysis}
\label{supp:ablation_stats}

\begin{figure}[H]
    \centering
    \includegraphics[width=1\textwidth]{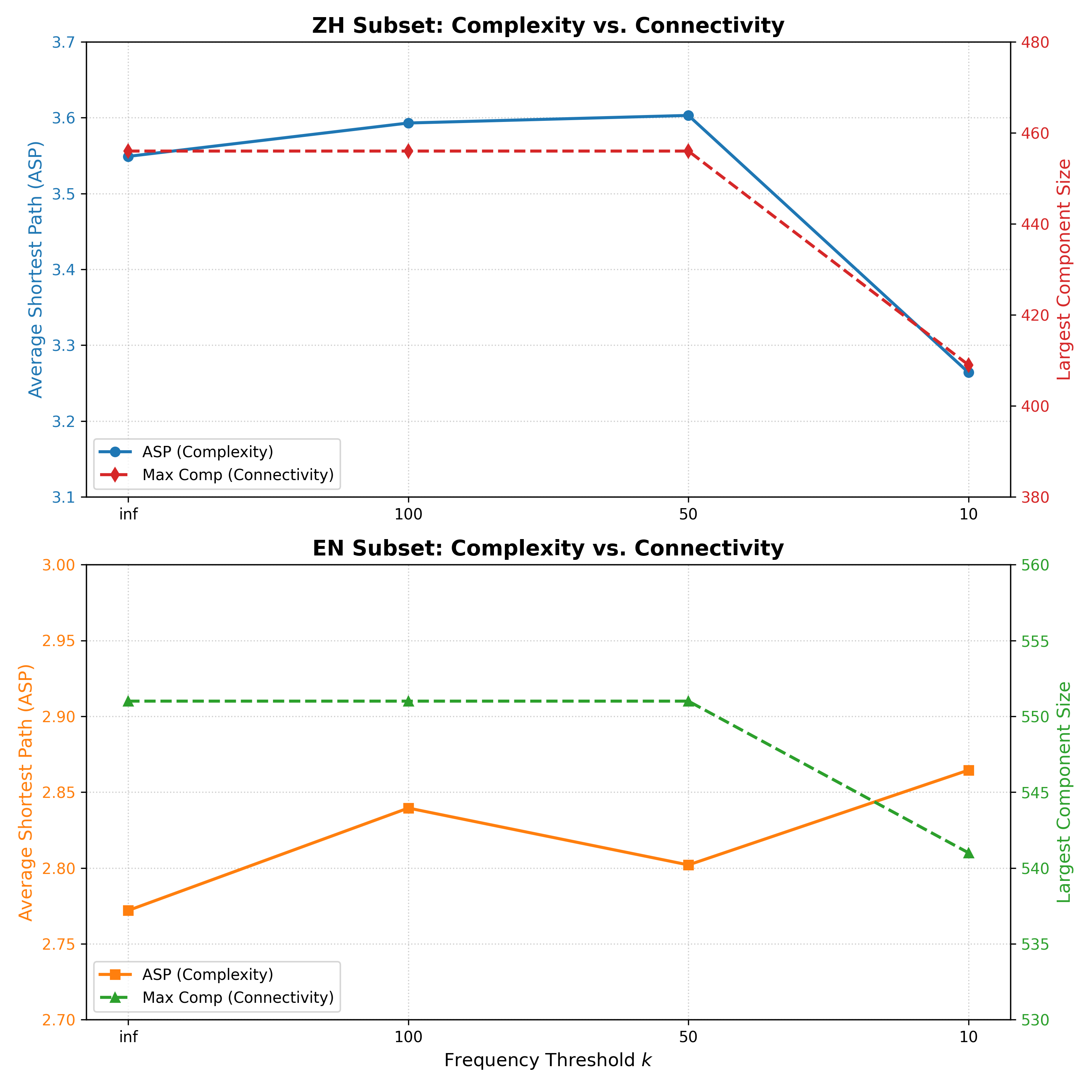}
    \caption{Bilingual ablation study of frequency threshold $k$ on Knowledge Graph topology. The dual-axis plots contrast inferential complexity (ASP, left axis) with structural connectivity (Largest Component Size, right axis).}
    \label{fig:k_ablation_combined}
\end{figure}

The choice of a unified threshold $k=50$ is governed by the need to balance maximum reasoning complexity with the structural integrity of the clinical cascades across both languages. As illustrated in Figure \ref{fig:k_ablation_combined}, the Chinese subset reaches its peak complexity with an Average Shortest Path (ASP) of 3.603 at $k=50$ while maintaining a perfectly stable largest component size of 456 nodes, indicating that $k=50$ surgically prunes generic hubs without inducing graph fragmentation. However, reducing $k$ further to 10 triggers a simultaneous collapse in both complexity and connectivity as the graph shatters into isolated fragments. For the English subset, although $k=100$ shows a localized peak, we enforce the unified threshold of $k=50$ to maintain a standardized evaluation protocol and ensure structural consistency across the bilingual dataset. At $k=50$, the EN-KG remains highly robust with a stable component size of 551 nodes and an ASP of 2.802, which is significantly superior to the unshattered baseline of $d_{original}=2.772$. By aligning both languages at $k=50$, we establish a rigorous and consistent benchmark that maximizes the diagnostic challenge for the dominant Chinese hard vignettes while ensuring the English reasoning paths are physically regularized against topological shortcuts.

\newpage
\section{Dataset Quality Assessment Prompt}
\label{supp:dataset_prompt}

To automatically categorize the synthesized multi-hop QA pairs and rigorously evaluate their intrinsic quality, we employed an ensemble of strong frontier LLMs as expert adjudicators. This prompt was utilized to generate the clinical task taxonomy and the 1-5 quality scores (Clarity, Validity, Difficulty) reported in Section \ref{subsec:dataset_quality}. Figure \ref{fig:quality_prompt} presents the exact template.

\begin{figure}[H] 
    \centering
    \fbox{
    \begin{minipage}[t]{0.85\textwidth}
        \small
        \textbf{Dataset Quality and Taxonomy Prompt}\\
        \hrule \vspace{2mm}
        You are an expert medical adjudicator and taxonomist. Analyze the following medical QA pair and provide your assessment in strict JSON format. \\
        
        \textbf{Categories for 'clinical\_task':} \\
        1. Basic Medicine (mechanisms, pathophysiology, anatomy) \\
        2. Clinical Diagnosis (recognizing disease via symptoms, labs, imaging) \\
        3. Clinical Treatment (treatment plans, surgery, medications, prognosis) \\
        4. Pharmacy/Drug Safety (indications, adverse effects, interactions) \\
        5. Prevention/Epidemiology (risk factors, screening, public health) \\
        6. Medical Humanities (ethics, communication, consent) \\
        
        \textbf{Categories for 'reasoning\_type':} \\
        Fact Retrieval, Single-hop, Multi-hop, Conditional Logic \\
        
        \textbf{Evaluate these 3 quality metrics on a scale of 1 to 5 (1=Poor, 5=Excellent):} \\
        - 'clarity\_score': Is the question unambiguous and well-formulated? \\
        - 'validity\_score': Is the rationale medically sound and accurate? \\
        - 'difficulty\_score': Does this require deep clinical expertise? \\
        
        Output EXACTLY this JSON format, nothing else: \\
        \{ \\
        \hspace*{4mm} "clinical\_task": "Category Name", \\
        \hspace*{4mm} "reasoning\_type": "Type Name", \\
        \hspace*{4mm} "clarity\_score": 5, \\
        \hspace*{4mm} "validity\_score": 5, \\
        \hspace*{4mm} "difficulty\_score": 4 \\
        \}
    \end{minipage}
    }
    \caption{The prompt template used for automated taxonomy classification and dataset quality scoring.}
    \label{fig:quality_prompt}
\end{figure}

\newpage
\section{Fine-Grained Dataset Evaluation Breakdown}
\label{supp:dataset_breakdown}

Table \ref{tab:Supplementary_breakdown} presents the exhaustive, fine-grained breakdown of all individual subsets and language variants evaluated in Section \ref{sec:benchmark_comparison}. All datasets were processed using the identical zero-shot evaluation prompt detailed in Supplementary \ref{supp:dataset_prompt}.

\begin{table}[H] 
\caption{Complete fine-grained metrics for all evaluated medical QA benchmarks.}
\label{tab:Supplementary_breakdown}
\centering
\resizebox{\textwidth}{!}{
\begin{tabular}{@{}llcccc@{}}
\toprule
\textbf{Category} & \textbf{Dataset (Subset/Language)} & \textbf{Size ($N$)} & \textbf{Clarity} & \textbf{Validity} & \textbf{Difficulty} \\ \midrule

\multirow{3}{*}{\textbf{MedQA}} 
& Mainland Test (ZH) & 3,425 & 3.88 & 3.78 & 2.91 \\
& Taiwan Test (ZH) & 1,413 & 3.95 & 3.90 & 2.97 \\
& US Test (EN) & 1,272 & 3.99 & 4.10 & 3.06 \\ \midrule

\multirow{6}{*}{\textbf{MMedBench}} 
& Chinese Test (ZH) & 3,426 & 3.98 & 3.92 & 2.82 \\
& English Test (EN) & 1,273 & 4.07 & 4.06 & 3.00 \\
& French Test (FR) & 622 & 4.03 & 4.13 & 2.81 \\
& Japanese Test (JA) & 199 & 3.91 & 3.86 & 2.98 \\
& Russian Test (RU) & 256 & 4.07 & 4.18 & 2.48 \\
& Spanish Test (ES) & 2,742 & 4.09 & 4.15 & 2.66 \\ \midrule

\multirow{6}{*}{\textbf{MultiMedQA}} 
& Anatomy & 154 & 4.10 & 4.24 & 2.51 \\
& Clinical Knowledge & 299 & 3.94 & 3.86 & 2.50 \\
& College Biology & 165 & 3.70 & 3.62 & 2.55 \\
& College Medicine & 200 & 3.83 & 3.70 & 2.51 \\
& Medical Genetics & 116 & 3.90 & 3.98 & 2.55 \\
& Professional Medicine & 308 & 4.05 & 4.19 & 3.01 \\ \midrule

\multirow{2}{*}{\textbf{Other Baselines}} 
& AfrimedQA & 3,723 & 3.80 & 3.72 & 2.83 \\
& PubMedQA & 1,000 & 2.92 & 2.43 & 2.46 \\ \midrule

\multirow{3}{*}{\textbf{AI-Assisted}} 
& CRAFT-MedQA (L) & 5,000 & 3.71 & 3.52 & 2.55 \\
& MedReason & 25,484 & 3.79 & 3.81 & 3.04 \\
& MedQA-Evol (UltraMedical) & 51,809 & 4.17 & 4.32 & 2.94 \\ \midrule

\textbf{Ours} 
& \textbf{ShatterMed-QA (Overall)} & \textbf{10,558} & \textbf{4.70} & \textbf{4.87} & \textbf{3.11} \\ \bottomrule
\end{tabular}
}
\end{table}

\newpage
\section{Qualitative Analysis: Single-Hop Shortcuts vs. Multi-Hop Reasoning}
\label{supp:case_study_details}

\begin{figure}[H] 
    \centering
    \includegraphics[width=1\textwidth]{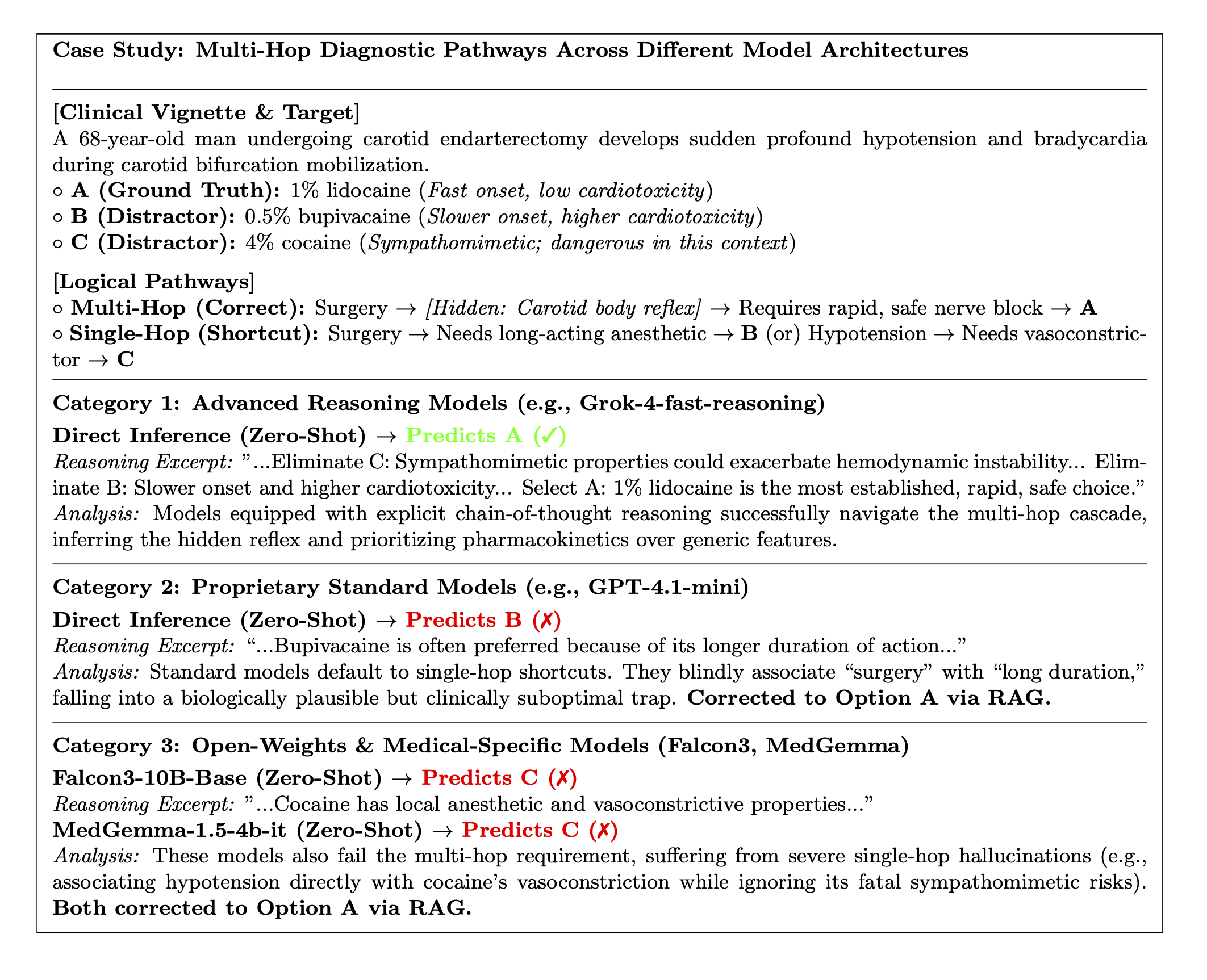} 
    \caption{Comparative analysis of model behaviors under the topological stress of ShatterMed-QA.}
    \label{fig:model_behavior_analysis}
\end{figure}

To concretely illustrate the operational difference between single-hop factual recall and multi-hop reasoning, Figure \ref{fig:model_behavior_analysis} demonstrates a clinical case where the implicit bridge entity (the carotid body reflex) is crucial for accurate diagnosis.

\textbf{Ground Truth Rationale:} During carotid endarterectomy, traction or manipulation of the carotid bifurcation may stimulate the baroreceptors in the carotid sinus. These receptors transmit signals via the Hering nerve to the nucleus tractus solitarius, increasing vagal activity and resulting in marked bradycardia and hypotension. Therefore, to minimize further hemodynamic responses during continued surgical manipulation, it is necessary to block the afferent sensory input from the carotid sinus. A common method to achieve this is local anesthetic infiltration around the carotid sheath. Among the options, lidocaine (Option A) has a rapid onset of action and good tissue penetration, making it particularly suitable for rapid intraoperative nerve blockade. In contrast, bupivacaine (Option B) has a slower onset, making it less useful when an immediate effect is required during surgery. 

\textbf{Shortcut Vulnerability:} Models relying on single-hop heuristics predictably fail against our topology-driven distractors. For instance, GPT-4.1-mini superficially links the keyword "surgery" to the longest-acting anesthetic (Option B), while others map "hypotension" directly to a vasoconstrictor (Option C), fatally ignoring systemic risks. Only models capable of advanced exclusionary reasoning successfully traverse the full cascade, inferring the hidden parasympathetic reflex to deduce the correct rapid-onset block (Option A). 

Crucially, these failures stem from internal parametric gaps rather than flawed question logic. As shown in Table \ref{supp:case_study_stats}, when the text snippet containing the missing bridge entity is randomly embedded among the top-$k$ ($k=5$) retrieved contexts via RAG, 16 out of 17 initially failing models successfully locate the evidence, resolve the multi-hop chain, and arrive at the ground truth.

\begin{table}[H] 
\caption{Performance shift of 21 evaluated models on the clinical case study. The widespread correction from incorrect (\ding{55}) to correct (\ding{51}) under RAG validates the logical integrity of the $k$-Shattered vignette.}
\label{supp:case_study_stats}
\centering
\footnotesize
\begin{tabular}{@{}llcc@{}}
\toprule
\textbf{Performance Shift} & \textbf{Model Name} & \textbf{Direct Answer} & \textbf{RAG Answer} \\ \midrule

\multirow{16}{*}{\textbf{\ding{55} $\rightarrow$ \ding{51} (Use RAG)}} 
& BioMistral-7B & E & A \\
& Falcon3-10B-Base & C & A \\
& Llama-3.1-8B & B & A \\
& Llama3-OpenBioLLM-8B & E & A \\
& Qwen3-14B-Base & B & A \\
& Yi-1.5-9B & B & A \\
& c4ai-command-r7b-12-2024 & C & A \\
& gemma-2-9b & C & A \\
& gpt-4.1-mini & B & A \\
& gpt-4.1-nano & C & A \\
& gpt-5-mini & C & A \\
& gpt-5-nano & C & A \\
& grok-4-1-fast-non-reasoning & C & A \\
& grok-4-fast-non-reasoning & B & A \\
& internlm3-8b-instruct & C & A \\
& medgemma-1.5-4b-it & C & A \\ \midrule

\multirow{4}{*}{\textbf{\ding{51} $\rightarrow$ \ding{51} (Correct Both)}} 
& Llama3-Med42-8B & A & A \\
& grok-4-1-fast-reasoning & A & A \\
& grok-4-fast-reasoning & A & A \\
& meditron-7b & A & A \\ \midrule

\textbf{\ding{55} $\rightarrow$ \ding{55} (Failed Both)} & granite-3.3-8b-base & B & B \\ \bottomrule
\end{tabular}
\end{table}

\newpage
\section{More Samples}

\begin{minipage}{\linewidth}
    \centering
    \fbox{
    \begin{minipage}{1\linewidth}
        \small
        \textbf{Case Study: Multi-Hop Diagnostic Pathways Across Different Model Architectures} \vspace{1mm} \\
        \hrule \vspace{2mm}
        
        \textbf{[Clinical Vignette \& Target]} \\
        A 25-year-old man presents with lifelong productive cough, recurrent sinusitis, reduced fertility, dextrocardia, and lower lobe bronchiectasis. He asks about risks following viral upper respiratory infections. Damage to which airway defense predisposes him to secondary bacterial pneumonia, and what is the most common causative pathogen? \\
        $\circ$ \textbf{A (Ground Truth):} \textit{Streptococcus pneumoniae} (\textit{Most common overall post-viral pathogen}) \\
        $\circ$ \textbf{B (Distractor):} \textit{Staphylococcus aureus} (\textit{Classic exam ''buzzword'' for post-influenza, but less common overall}) \\
        $\circ$ \textbf{C (Distractor):} \textit{Haemophilus influenzae} (\textit{Common in chronic bronchiectasis, but misses the ''post-viral'' context}) \vspace{2mm}

        \textbf{[Logical Pathways]} \\
        $\circ$ \textbf{Multi-Hop (Correct):} Vignette $\rightarrow$ \textit{[Hidden: Kartagener Syndrome / Impaired Mucociliary Clearance]} $\rightarrow$ Viral URI further damages cilia $\rightarrow$ Most common secondary pathogen overall $\rightarrow$ \textbf{A} \\
        $\circ$ \textbf{Single-Hop (Shortcut 1):} Bronchiectasis $\rightarrow$ Exacerbation pathogen $\rightarrow$ \textbf{C} \\
        $\circ$ \textbf{Single-Hop (Shortcut 2):} Post-viral pneumonia $\rightarrow$ Classic medical board trope $\rightarrow$ \textbf{B} \vspace{2mm}
        
        \hrule \vspace{2mm}
        \textbf{Category 1: The ''Post-Viral Buzzword'' Shortcut (e.g., GPT-5-mini, GPT-4.1-mini)} \vspace{1mm} \\
        \textbf{Direct Inference (Zero-Shot) $\rightarrow$ \textcolor{red}{Predicts B (\ding{55})}} \\
        \textit{Reasoning Excerpt:} ''...features of primary ciliary dyskinesia (Kartagener syndrome)... Viral upper respiratory infections damage mucociliary defenses and predispose to bacterial superinfection, classically by Staphylococcus aureus...'' \\
        \textit{Analysis:} Frontier models successfully navigate the first hop (diagnosing Kartagener syndrome) but fall into a statistical shortcut on the second hop. They overfit to the classic ''exam trope'' linking post-viral pneumonia directly to \textit{S. aureus}, ignoring the epidemiological reality that \textit{S. pneumoniae} remains the most common overall. \textbf{Corrected to Option A via RAG.} \vspace{2mm}

        \hrule \vspace{2mm}
        \textbf{Category 2: The ''Bronchiectasis'' Shortcut (e.g., Grok-4-fast-reasoning, Qwen3-14B, MedGemma)} \vspace{1mm} \\
        \textbf{Direct Inference (Zero-Shot) $\rightarrow$ \textcolor{red}{Predicts C (\ding{55})}} \\
        \textit{Reasoning Excerpt:} ''...Patient Diagnosis: Kartagener syndrome... lower lobe bronchiectasis... Haemophilus influenzae is the most common colonizer and cause of exacerbations in bronchiectasis...'' \\
        \textit{Analysis:} These models anchor heavily on the ''bronchiectasis'' keyword, triggering a superficial single-hop association with \textit{H. influenzae}. They completely bypass the critical conditional logic required by the prompt: identifying the pathogen specifically subsequent to an \textit{acute viral infection}. \textbf{Corrected to Option A via RAG.} \vspace{2mm}

        \hrule \vspace{2mm}
        \textbf{Category 3: Rigid Parametric Failure (e.g., Llama3-Med42-8B)} \vspace{1mm} \\
        \textbf{Direct Inference (Zero-Shot) $\rightarrow$ \textcolor{red}{Predicts C (\ding{55})}} \quad | \quad \textbf{RAG Integration $\rightarrow$ \textcolor{red}{Predicts C (\ding{55})}} \\
        \textit{Analysis:} Certain domain-specific models fail both the zero-shot and RAG settings. Despite the RAG context explicitly stating that \textit{S. pneumoniae} is the most common pathogen in this setting, the model's internal parametric bias towards the bronchiectasis-\textit{H. influenzae} shortcut is too rigid to be overridden by external multi-hop evidence. \vspace{1mm}
        
    \end{minipage}
    }
    \vspace{2mm}
    \captionof{figure}{Comparative analysis of model behaviors under the topological stress of ShatterMed-QA (Kartagener Syndrome and Secondary Pneumonia).}
    \label{fig:model_behavior_analysis_2}
\end{minipage}

\begin{figure}[H]
    \centering
    \fbox{
    \begin{minipage}{1\linewidth}
        \small
        \textbf{Case Study: Temporal Pharmacodynamics and Semantic Traps in Endocrine Pathways} \vspace{1mm} \\
        \hrule \vspace{2mm}
        
        \textbf{[Clinical Vignette \& Target]} \\
        A 38-year-old woman with symptomatic uterine leiomyomas is started on a 3-month depot injectable that provides "continuous stimulation of pituitary gonadotrophs" to reduce uterine size preoperatively. Eight weeks later, she experiences amenorrhea and hot flashes. Which pattern of circulating pituitary gonadotropins is most likely seen at this time? \\
        $\circ$ \textbf{A (Ground Truth):} Low LH, low FSH (\textit{Receptor downregulation after initial flare}) \\
        $\circ$ \textbf{B/C (Distractors):} High LH, low FSH / High FSH, low LH (\textit{Asymmetric suppression; physiologically inconsistent here}) \\
        $\circ$ \textbf{D (Distractor):} High LH, high FSH (\textit{Semantic trap; reflects the initial flare phase, not the 8-week mark}) \vspace{2mm}

        \textbf{[Logical Pathways]} \\
        $\circ$ \textbf{Multi-Hop (Correct):} "Continuous stimulation" depot $\rightarrow$ GnRH agonist $\rightarrow$ Initial flare $\rightarrow$ \textit{[Hidden temporal shift: prolonged exposure > 2 weeks causes receptor downregulation]} $\rightarrow$ Suppressed pituitary axis $\rightarrow$ \textbf{A} \\
        $\circ$ \textbf{Single-Hop (Shortcut):} Reads "continuous stimulation" $\rightarrow$ Blindly assumes hypersecretion $\rightarrow$ \textbf{D} \vspace{2mm}
        
        \hrule \vspace{2mm}
        \textbf{Category 1: Advanced Temporal Reasoning (e.g., GPT-4.1, Grok-4-fast-reasoning)} \vspace{1mm} \\
        \textbf{Direct Inference (Zero-Shot) $\rightarrow$ \textcolor{green}{Predicts A (\ding{51})}} \\
        \textit{Reasoning Excerpt:} "...Continuous stimulation of pituitary gonadotrophs by a long-acting GnRH agonist initially causes a transient surge of LH and FSH, but then leads to receptor downregulation... Therefore both LH and FSH will be low during the maintenance phase (8 weeks)..." \\
        \textit{Analysis:} Frontier models successfully navigate the temporal pharmacodynamics. They recognize that the literal text ("continuous stimulation") results in a paradoxical physiological outcome (downregulation) due to the specific 8-week timeline provided in the vignette. \vspace{2mm}

        \hrule \vspace{2mm}
        \textbf{Category 2: The ''Semantic Trap'' Shortcut (e.g., Gemma-2-9b, BioMistral-7B, c4ai-command)} \vspace{1mm} \\
        \textbf{Direct Inference (Zero-Shot) $\rightarrow$ \textcolor{red}{Predicts D (\ding{55})}} \\
        \textit{Reasoning Excerpt:} "...The patient is receiving a depot injectable that provides continuous stimulation of pituitary gonadotrophs. This will result in high levels of both LH and FSH in the circulation..." \\
        \textit{Analysis:} These models fall victim to a direct semantic shortcut. They anchor on the phrase "continuous stimulation" and superficially map it to "high hormone levels" (Option D), completely bypassing the crucial multi-hop deduction required for GnRH agonist receptor downregulation. \textbf{Corrected to Option A via RAG.} \vspace{2mm}

        \hrule \vspace{2mm}
        \textbf{Category 3: Rigid Parametric Failure (e.g., Granite-3.3-8b-base, Llama3-Med42-8B)} \vspace{1mm} \\
        \textbf{Direct Inference (Zero-Shot) $\rightarrow$ \textcolor{red}{Predicts D or B (\ding{55})}} \quad | \quad \textbf{RAG Integration $\rightarrow$ \textcolor{red}{Predicts B (\ding{55})}} \\
        \textit{Analysis:} Certain models fail to synthesize the correct physiological mechanism even when external RAG evidence hints at the hypoestrogenic state (amenorrhea/hot flashes) induced by the therapy. Their internal biases toward the literal "stimulation" keyword are too rigid to overcome, resulting in persistent failures. \vspace{1mm}
        
    \end{minipage}
    }
    \vspace{2mm}
    \caption{Comparative analysis of model behaviors under the topological stress of ShatterMed-QA (GnRH Agonist Pharmacodynamics).}
    \label{fig:endocrine_case_study}
\end{figure}

For a broader selection of multi-hop clinical case studies across both English and Chinese splits, please visit our interactive showcase at \url{https://shattermed-qa-web.vercel.app/#showcase}.

\end{document}